# Deep Transfer Learning for Error Decoding from Non-Invasive EEG

Martin Völker[1,2], Robin T. Schirrmeister[1,2], Lukas D. J. Fiederer[1], Wolfram Burgard[2], Tonio Ball[1]

[1] Translational Neurotechnology Lab, University Medical Center Freiburg, Germany

[2] Department of Computer Science, Alberts-Ludwigs-University Freiburg, Germany

martin.voelker@uniklinik-freiburg.de

*Abstract—* We recorded high-density EEG in a flanker task experiment (31 subjects) and an online BCI control paradigm (4 subjects). On these datasets, we evaluated the use of transfer learning for error decoding with deep convolutional neural networks (deep ConvNets). In comparison with a regularized linear discriminant analysis (rLDA) classifier, ConvNets were significantly better in both intra- and inter-subject decoding, achieving an average accuracy of 84.1 % within subject and 81.7 % on unknown subjects (flanker task). Neither method was, however, able to generalize reliably between paradigms. Visualization of features the ConvNets learned from the data showed plausible patterns of brain activity, revealing both similarities and differences between the different kinds of errors. Our findings indicate that deep learning techniques are useful to infer information about the correctness of action in BCI applications, particularly for the transfer of pre-trained classifiers to new recording sessions or subjects.

## I. INTRODUCTION

The classification and subsequent correction of errors has become a topic of great interest in the field of brain-computer interfacing (BCI), as most existing systems are still too error-prone for real-life application. To further enhance practicality, error decoding has been implemented in a number of BCI applications in the last years, e.g., P300 spellers [1], shared-control BCIs [2], observation of complex robot actions [3], or high-precision motor tasks involving intracranial EEG [4,5].

Another obstacle preventing practical real-life use of BCIs is that in many cases large amounts of training data are required, acquisition of which often needs to be repeated even for new sessions of the same subject. Progress in transfer learning could help to solve this problem: it could reduce overall training time, and enable using pre-trained classifiers on new subjects, without additional training. For the latter, it is important to find out how many training subjects such a classifier needs to reliably generalize to new subjects.

One class of methods that has been successfully used for transfer learning tasks are deep learning techniques, which in the past years have revolutionized fields of research including speech recognition and computer vision. In the field of BCI research, it has recently been shown that deep convolutional neural networks (deep ConvNets) are able to compete with state-of-the-art methods in decoding of hand and foot movements [6] or P300 signals [7] from EEG data. There is also great interest in the interpretability and visualization of deep neural networks [8]. To date, ConvNets have however not been evaluated with respect to error decoding.

In this study, we analyzed the performance of deep ConvNets regarding decoding of errors from non-invasive EEG in two paradigms (flanker task with 31 subjects, online GUI control with 4 subjects). We analyzed transfer learning across different recording days, across different subjects, and across two different paradigms, and compared the ConvNet results to those obtained by regularized linear discriminant analysis (rLDA). We could show that deep ConvNets performed significantly better than rLDA in intra-subject decoding and in transferring to unknown subjects. In the flanker task, average deep ConvNet error decoding accuracies were also higher than accuracies previously reported in literature.

## II. RELATED WORK

A selection of representative studies on intra-subject error decoding from 2008-2016 is listed in Table 1.

TABLE I. STUDIES ON INTRA-SUBJECT ERROR DECODING

| source & year | normalized accuracy (%) | channels used for decoding | number of subjects | decoding method | features |
|---|---|---|---|---|---|
| [9] 2008 | 82.5 | 2 | 5 | Gaussian classifier | Voltage (<10Hz) |
| [10] 2009 | 74 | 2 | 13 | LDA & Gaussian classifier | Voltage (<10Hz) |
| [11] 2012 | 70 | 1 | 7 | Logistic regression | Voltage (<10Hz) |
| [2] 2013 | 74.4 | 8 | 4 | rLDA | Voltage (<10Hz) |
| [12] 2015 | 71.1 | 8 | 6 | rLDA | Voltage (<10Hz) |
| [13] 2015 | 77 | 32 | 8 | SVM | CSP-Voltage (<10Hz) + Power (4-8Hz) |
| [14] 2016 | 74 | 8 | 4 | LDA | Voltage (<10Hz) |

When one compares classification accuracies in error decoding, it is important to use the normalized recording accuracy, i.e., the arithmetic mean of the single class decoding accuracies. In that way, the typically-present trial imbalance (more correct than error trials) does not distort the accuracy values, as the 50% chance level is maintained.

In most previous studies, the reported normalized accuracies lay between 70 and 80%. Popular decoding methods were Gaussian classifiers, logistic regression, (regularized) linear discriminant analysis ((r)LDA), and support vector machines (SVM), sometimes combined with common spatial pattern (CSP) feature extraction. There are only few data about error decoding with pre-trained classifiers on unknown subjects [15].

## III. EXPERIMENTS

We tested deep learning for error decoding on the following two paradigms:

### A. Eriksen Flanker Task

We recorded 128-channel high-density EEG from 31 healthy subjects instructed to perform a flanker task (Fig.1) within an electro-magnetically shielded environment.

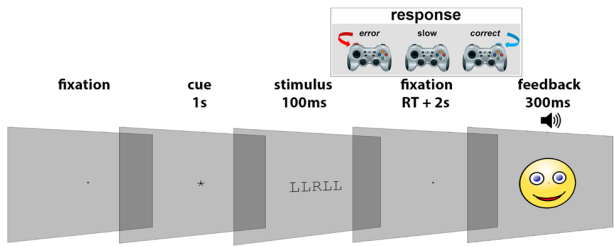

Figure 1. Flanker task paradigm. Subjects had to react quickly to the central letter of a 5-letter stimulus by pressing a button with either their left ('L') or right ('R') index finger. Each experiment consisted of 1000 trials [16].

The subjects had to act under time pressure, being required to react within an individual reaction time limit which was set to the individual mean reaction time in a training phase. An "error" in this paradigm meant that the subject reacted with the wrong hand to the middle letter (R or L) of the stimulus. On average, the subjects had an error rate of 22.2 ± 1.8 % (mean ± standard error of the mean (SEM)).

### B. Online GUI for the control of intelligent robots

64-channel EEG was recorded in an online BCI setup, in which 4 subjects controlled a mobile robot by issuing high-level commands through a graphical user interface (GUI) and as described in [17]. The control of the GUI was achieved with 4 different mental tasks, decoded in real time by an adaptive ConvNet classifier. Fig. 2 illustrates the course of a typical trial. An "error" consisted of a wrong step in the GUI control given the intended aim of the subject.

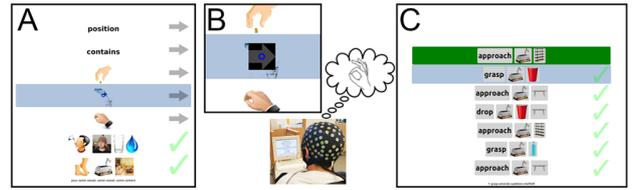

Figure 2. Online GUI control. **A)** The planner interface shows a selection of possible commands. **B)** A cue for 1 of 4 possible commands appears (here: right hand movement imagination for the "selection" command); the subject starts the mental task. **C)** When the online classifier predictions exceed the significance threshold, the decoded command is executed in the GUI.

Note that in session 0, the subjects performed the same tasks as in the online sessions, but were in fact not in control of the GUI, and errors were randomly introduced with about 20% error rate. Starting with session 1, the subjects controlled the GUI with an online deep learning classifier which was trained on the first session and continuously adapted with past trials. On average, each subject completed 3032 ± 818 trials and had an error rate of 25.3 ± 1.9 %. Also note that errors were not decoded online; here we report *post hoc* error decoding results based on the online experiment data.

## IV. PREPROCESSING AND CLASSIFIER DESIGN

The EEG data were re-referenced to a common average reference (CAR) and resampled to 250 Hz. In the case of the deep ConvNets, an electrode-wise exponential running standardization [6] was applied with a decay factor of 0.999; the rLDA was more accurate without the running standardization and was thus applied without it. As classifier input, the data was epoched from 500 ms before the response or GUI event until 1000 ms after it.

For both rLDA and ConvNet classifiers, python implementations were used. To estimate the shrinkage parameter for the rLDA, we used scikit-learn's Ledoit-Wolf estimator [18]. Deep ConvNets were designed using the open-source braindecode toolbox (v0.3.0), with the same architecture as in [6]. A stride of 2 samples was used to create a smaller receptive field without changing the number of layers.

## V. RLDA OPTIMIZATION

To compare deep ConvNets to a strong alternative, we optimized the rLDA procedure beforehand on data of the flanker task experiment, leading to several observations. First, applying an exponential running standardization on the data reduced the normalized decoding accuracy of the rLDA. Second, a selection of seven midline electrodes yielded better accuracies than the whole 128-channel set for rLDA. Third, between-subject decoding on 128 electrodes worked best for lower sampling rates, and dropped in accuracy when higher sampling rates were used (Fig. 3). However, the midline channel selection gained in accuracy at higher sampling rates. Therefore, we used the midline channel selection with a high sampling rate as comparison to the deep ConvNet decoding results.

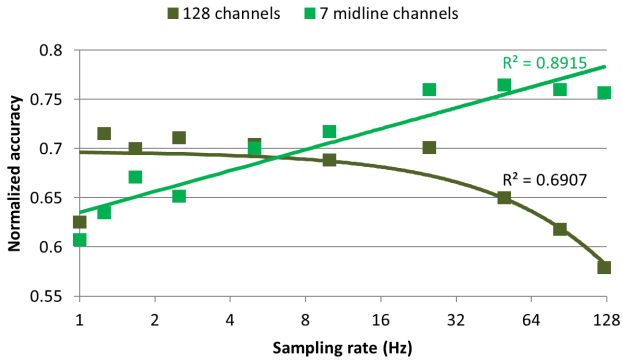

Figure 3. Influence of channel selection and sampling rate on between-subject error decoding using an rLDA classifier. With features of 128 EEG channels (dark green, logarithmic fit), resampling to lower sampling rates resulted in better decoding accuracies, with a maximum of 71.48 % at 1.25 Hz. Voltage features of selected channels (light green, exponential fit), yielded the best mean decoding accuracies with higher sampling rates peaking at 50 Hz with 76.45 %. The 7 midline channels were Cz, CPz, FCz, Fz, Pz, POz, and Fpz.

## VI. WITHIN-SUBJECT DECODING

### A. Comparison of decoding methods

Previously, it was reported that an rLDA classifier worked best for error decoding on a flanker task data set and outperformed other state-of-the-art methods in a within-subject cross-validation [13]. Here we show that deep ConvNets achieved significantly higher intra-subject decoding accuracies than rLDA, averaged over all 31 subjects included in our study (Fig. 4A). In the online GUI experiment, the average over 4 subjects followed the same trend (Fig. 4B).

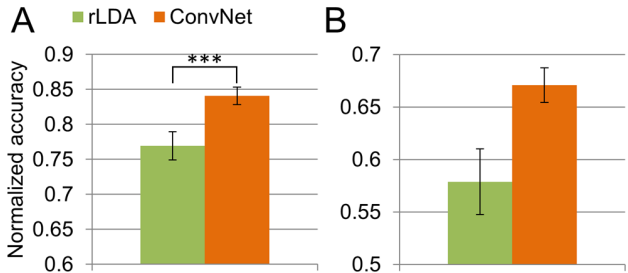

Figure 4. Comparison of within-subject decoding by rLDA and deep ConvNets. Error bars show the SEM. **A)** Flanker task (mean of 31 subjects), last 20% of subject data as test set. Deep ConvNets were 7.12% better than rLDA, pval = 6.24 *$10^{-20}$ (paired t-test). **B)** Online GUI control (mean of 4 subjects), last session of each subject as test data.

Note that the classifiers in Fig. 4B were tested on independent sessions of the same subjects, so that differences between recording days likely made it harder for the classifiers to decode errors reliably.

### B. Influence of subjective error recognition

In the online GUI control paradigm, the subjects were asked to count errors, i.e. when the executed action within the GUI did not match the mental task they were doing. The subjective error recognition rate is plotted in Fig. 5 together with the online GUI control accuracy and the error decoding accuracies of rLDA and ConvNets.

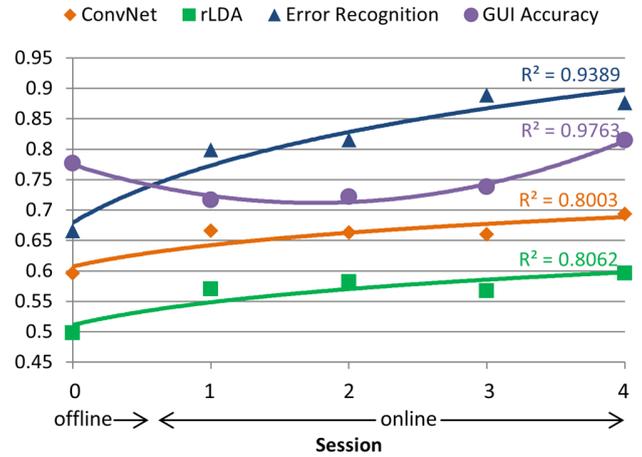

Figure 5. Influence of of error recognition rate on decoding performance. Average of 4 subjects of the Online GUI control paradigm (last offline session and first 4 online adaptive sessions). Accuracies of ConvNet (orange, logarithmic fit) and rLDA (green, logarithmic fit) were calculated with a leave-one-session-out cross-validation. The ratio of errors recognized by the subjects is plotted in blue (logarithmic fit), and the GUI accuracy is visualized in purple (polynomial fit $2^{nd}$ order).

The error recognition rate was positively correlated with session-wise cross-validated accuracies of rLDA ($r^2$ = 0.809) and ConvNets ($r^2$ = 0.810), while the GUI control accuracy showed no such correlations ($r^2$ = 0.005 and 0.001, respectively).

## VII. VISUALIZATION OF LEARNED ATTRIBUTES

We visualized the learned attributes of the networks by calculating input-perturbation network-prediction correlation maps [6]. Correlations of amplitude or frequency power changes with ConvNet decoding predictions were calculated by perturbing the input amplitudes either in the time domain or with respect to the spectral power in different frequency ranges and by computing the correlation of ConvNet outputs (last layer before softmax layer) with the input perturbations.

### A. Voltage features

Fig. 6 shows the average time-resolved input-perturbation network-prediction correlation maps for voltage features in both flanker task (A) and GUI control (B) paradigms. Note that features were only defined for visualization, while networks were trained in an end-to-end manner, i.e., using only minimally preprocessed raw EEG data (see classifier design) and without hand-engineered feature extraction.

Both in flanker task and GUI control, the networks' maximal correlations occurred 300 to 500 ms after the error. At this time, the networks seem to strongly rely on positive voltage changes at central to fronto-central electrodes to decode errors (Fig. 6, arrows). This time range is associated with the Pe component of the error response [19]. In the flanker task, this positive frontal midline correlation was slightly more anterior than in the GUI control experiment.

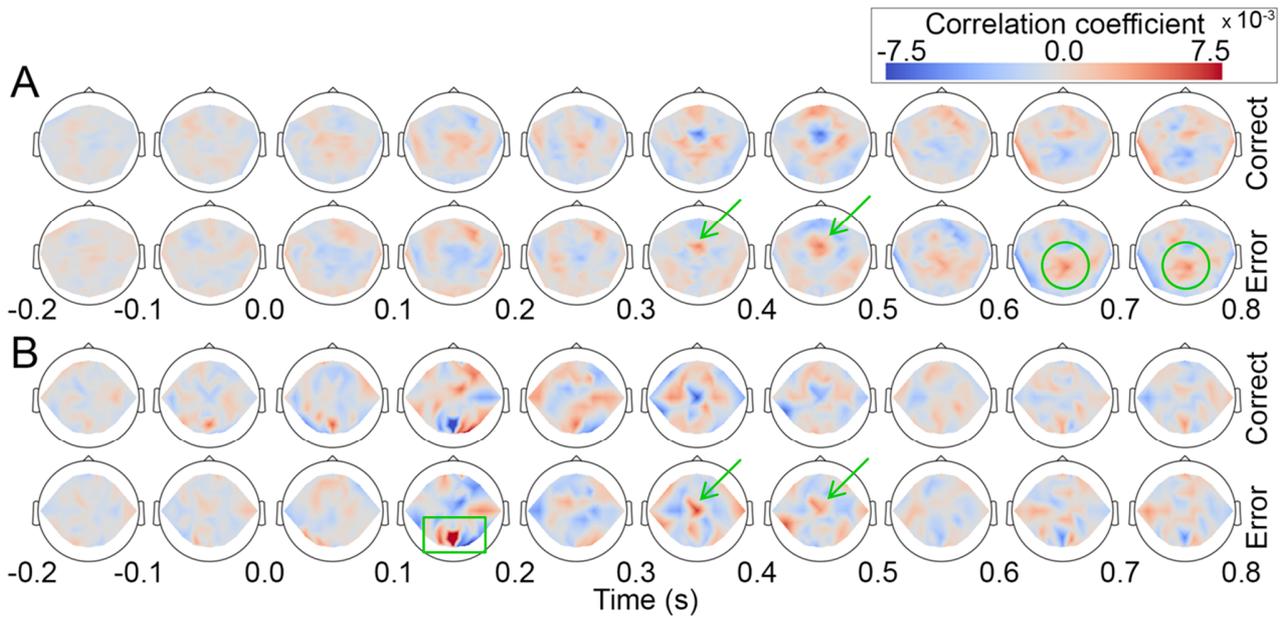

Figure 6. Time-resolved voltage feature input-perturbation network-prediction correlation maps, mean of 30 iterations. **A)** Flanker task, mean of 31 subjects (128 electrodes). **B)** GUI control, mean of 4 subjects (64 electrodes).

Beyond similarities, the correlation maps also suggested differences in the learned features in both paradigms. On the one hand, the ConvNets for the flanker task paradigm appeared to rely on more widespread, mid-parietal voltage changes between 600 and 800 ms (Fig. 6A, circles), which was less clear in the GUI control paradigm. On the other hand, the networks in the GUI control exhibited a positive occipital correlation peak shortly after the error event (Fig. 6B, rectangle), which may reflect the fact that error recognition in this paradigm was purely visual-input-based.

*B. Spectral power features*

In Fig. 7, input-perturbation network-prediction correlation maps are plotted in a frequency-resolved manner. As for the time-resolved maps, these maps revealed both similarities and differences between the features used by the trained ConvNets

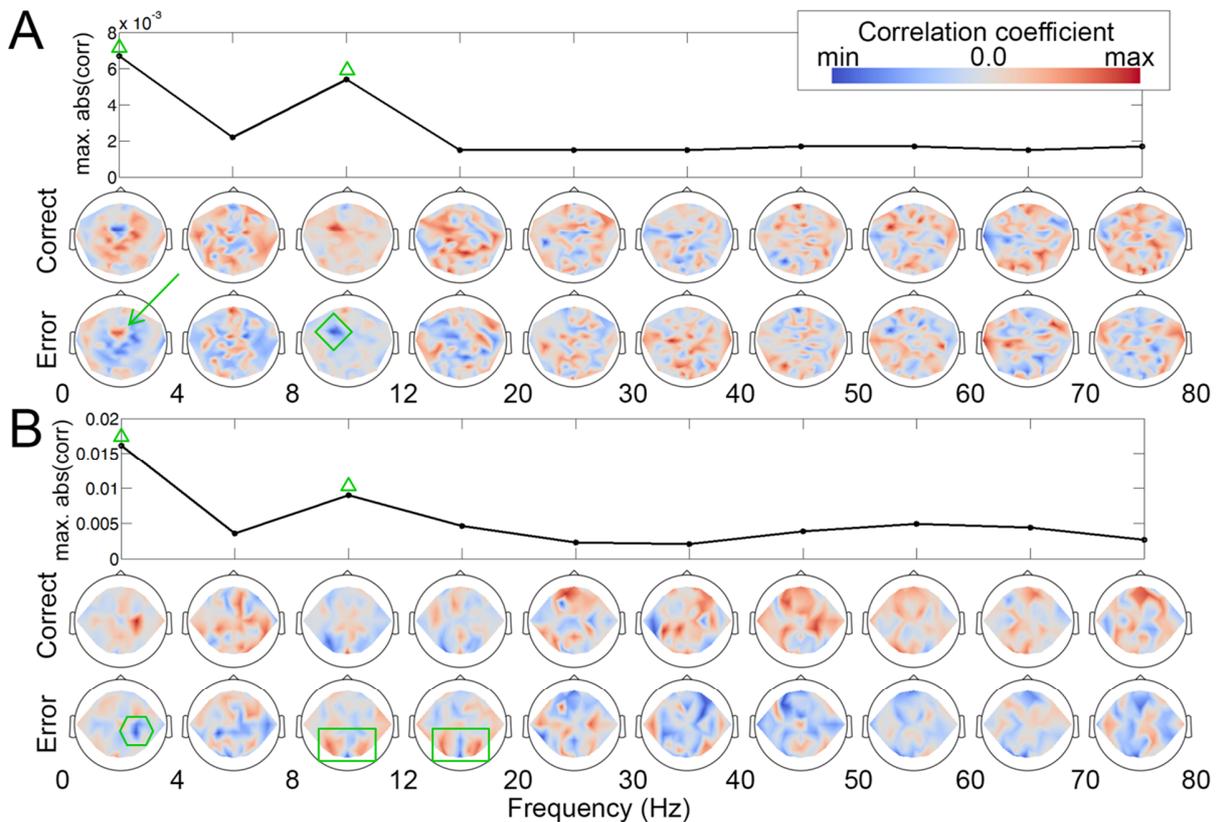

Figure 7. Frequency-resolved spectral power feature input-perturbation network-prediction correlation maps, mean of 30 iterations, for the time between 50 and 750 ms after the error. Above the topographical maps with individual colormap scaling, the maximum correlation of the respective frequency range is plotted. **A)** Flanker task, mean of 31 subjects (128 electrodes) with the deep ConvNet. **B)** GUI control, mean of 4 subjects (64 electrodes).

in the two experiments.

In both, the networks relied mostly on lower-frequency components, with a peak in delta and alpha bands (Fig. 7, triangles). However, other patterns of network-prediction correlations differed strongly between the paradigms. For example, a distinct positive error-related correlation peak at fronto-central channels in the delta range occurred in the flanker task (Fig. 7A, arrow), while there was a negative error-related correlation peak in the area of the right motor cortex in the GUI control paradigm (Fig. 7B, hexagon). In the alpha (8 - 12 Hz) range, prediction of the error class correlated strongly with perturbations consisting in a power decrease at left fronto-lateral areas in the flanker task (diamond), while the error class in the GUI control correlated with an occipital power decrease in the alpha and low-beta range and a power increase in the low-gamma range (rectangles), in the same region as the correlation with occipital voltage deflections (Fig. 6B, rectangles).

## VIII. BETWEEN-SUBJECT TRANSFER LEARNING

We evaluated between-subject decoding with a leave-one-subject-out cross-validation.

### A. Comparison of rLDA and deep ConvNets

We again compared rLDA and deep ConvNets for their decoding performance in new subjects (Fig. 8).

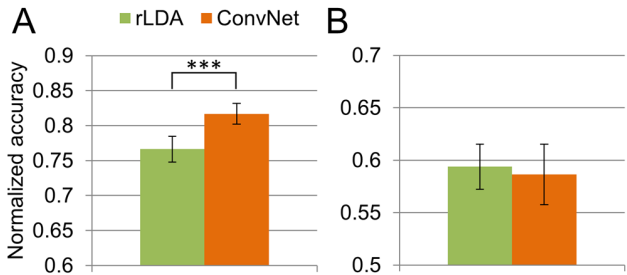

Figure 8. Mean normalized decoding accuracy on unknown subjects. Error bars show the SEM. **A)** Flanker task, trained on 30 subjects, tested on 1 subject. Deep ConvNets were 5.05% better than rLDA, p = 3.16 *$10^{-4}$ (paired t-test). **B)** Online GUI control. Trained on 3 subjects, tested on the respective remaining subject.

Deep ConvNets were significantly better in generalizing to new subjects in the flanker task. Remarkably, the deep ConvNets achieved 81.7 % normalized accuracy on new subjects, when trained on 30 other subjects. The comparatively bad performance of both classifiers on new subjects in the online GUI control likely results from the very small subject group to train on, compared to the flanker task (3 vs. 30 training subjects, also see next paragraph).

### B. Influence of training group size

In the design of BCI studies, an important point to consider is how many subjects a classifier might need to generalize to new subjects. We tested this issue on the flanker task dataset with a leave-one-subject out cross-validation with a varying number of subjects to train on (Fig. 9).

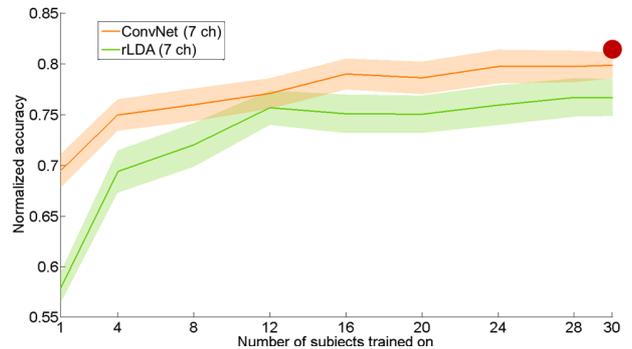

Figure 9. Influence of training group size on decoding performance in unknown subjects (flanker task). SEM is plotted as semi-transparent area. The red circle symbolizes the best between-subject decoding acuracy, achieved with ConvNet decoding from all 128 channels.

This evaluation showed that generalization of error decoding to new subjects was not reliable with small groups under 5 subjects. The accuracies for both rLDA and ConvNets still increased strongly up to 12-16 subjects to train on; more subjects improved the accuracy only slightly. ConvNets performed consistently better than the rLDA.

## IX. TRANSFER ACROSS PARADIGMS

We further tested the classifiers' abilities with using the whole dataset of one paradigm as training data, and testing on the whole dataset of the respective other paradigm (Table 2).

TABLE II. SUBJECT-AVERAGED BETWEEN-PARADIGM DECODING

| Method | Between-paradigm decoding accuracy ± SEM | |
|---|---|---|
| | *Flanker task to online GUI* | *Online GUI to flanker task* |
| 7ch rLDA | 0.5000 ± 0.0000 | 0.4993 ± 0.0026 |
| 7ch ConvNet | 0.4874 ± 0.0084 | 0.4244 ± 0.0116 |
| 64ch ConvNet | 0.4814 ± 0.0115 | 0.4248 ± 0.0310 |

Neither rLDA nor ConvNets was able to predict errors across paradigms. While the rLDA classifier in most cases predicted all trials for one of the two classes, the ConvNets had a bias towards the contrary class, resulting in an average decoding accuracy below chance level.

## X. CONCLUSION & OUTLOOK

Deep convolutional neural networks proved to be a well-working method for error decoding, and were significantly better than rLDA for within-subject error decoding in a flanker task as well as in *post hoc* analysis of errors that occurred during an online BCI GUI control experiment. In the flanker task, our deep ConvNets achieved the highest to date reported average accuracy. Also in contrast to some published studies with small (<10) numbers of subjects, we included 31 subjects in our flanker task experiment, which enabled robust investigation of between-subject transfer learning. Here we find that ConvNets were also significantly better than rLDA when applied to unknown subjects, setting a new benchmark with over 81 % normalized accuracy. For a generalization to new subjects, our data suggest that a training subject group of at least 15 subjects might be necessary for reliable error decoding on unknown subjects.

Visualization of learned signal features used by the deep learning model revealed physiologically plausible patterns. Time-resolved input-perturbation network-prediction correlation maps of both paradigms exhibited a similar effect at fronto-central channels 300 to 500 ms after an error (arrows in Fig. 6), i.e., above one of the main brain regions implicated in error processing. Patterns however also differed, e.g., in occipital areas. Frequency-resolved maps revealed more fine-grained patterns, which differed strongly for the two paradigms, possibly explaining the difficulty in inter-paradigm decoding. Further examination of the differences between motor execution errors and sensory perceived errors could help in the understanding of these effects.

As a next step, techniques including data augmentation and automated hyper-parameter and architecture search might help to improve the generalization of deep ConvNets, possibly also enabling transfer learning between different paradigms. Furthermore, an inclusion of more, different error-inducing paradigms could help to train a general error-recognition network for BCI applications.

ACKNOWLEDGMENT

This work was supported by DFG grant EXC1086 BrainLinks-BrainTools, Baden-Württemberg Stiftung grant BMI-Bot, Graduate School of Robotics in Freiburg, Germany, and the State Graduate Funding Program of Baden-Württemberg, Germany.

REFERENCES

[1] M. Spüler, M. Bensch, S. Kleih, W. Rosenstiel, M. Bogdan, & A. Kübler, "Online use of error-related potentials in healthy users and people with severe motor impairment increases performance of a P300-BCI." Clinical Neurophysiology 123.7 (2012): 1328-1337.

[2] I. Iturrate, L. Montesano, & J. Minguez, "Shared-control brain-computer interface for a two dimensional reaching task using EEG error-related potentials." In Engineering in Medicine and Biology Society (EMBC), 2013 35th Annual International Conference of the IEEE (pp. 5258-5262). IEEE.

[3] A.F. Salazar-Gomez, J. DelPreto, S. Gil, F.H. Guenther., & D. Rus, "Correcting robot mistakes in real time using eeg signals". ICRA 2017. IEEE.

[4] T. Milekovic, T. Ball, A. Schulze-Bonhage, A. Aertsen, & C. Mehring, "Detection of error related neuronal responses recorded by electrocorticography in humans during continuous movements." PloS one. 2013 Feb 1;8(2):e55235.

[5] N. Even-Chen, S.D. Stavisky, C. Pandarinath, P. Nuyujukian, C.H. Blabe, L.R. Hochberg, J.M. Henderson, & K.V. Shenoy, "Feasibility of Automatic Error Detect-and-undo system in Human Intracortical Brain-Computer Interfaces." IEEE Transactions on Biomedical Engineering. 2017 Nov 21.

[6] R.T. Schirrmeister, J.T. Springenberg, L.D.J. Fiederer, M. Glasstetter, K. Eggensperger, M. Tangermann, F. Hutter, W. Burgard, & T. Ball, "Deep learning with convolutional neural networks for EEG decoding and visualization." Human brain mapping (2017).

[7] R.K. Maddula, J. Stivers, M. Mousavi, S. Ravindran, & V.R. de Sa, "Deep recurrent convolutional neural networks for classifying P300 BCI signals." In Proceedings of the Graz BCI Conference 2017.

[8] I. Sturm, S. Lapuschkin, W. Samek, & K.R. Müller, "Interpretable deep neural networks for single-trial EEG classification." Journal of neuroscience methods (2016), 274, 141-145.

[9] P.W. Ferrez, & J.D.R. Millán, "Error-related EEG potentials generated during simulated brain–computer interaction." IEEE transactions on biomedical engineering, 2008, 55(3), 923-929.

[10] A. Kreilinger, C. Neuper, G. Pfurtscheller, & G.R. Müller-Putz., "Implementation of error detection into the graz-brain-computer interface, the interaction error potential." In European Conference for the Advancement of Assistive Technology 2009 (pp. 195-199).

[11] C. Vi, & S. Subramanian, "Detecting error-related negativity for interaction design." In Proceedings of the SIGCHI Conference on Human Factors in Computing Systems 2012 (pp. 493-502). ACM.

[12] R. Chavarriaga, I. Iturrate, Q. Wannebroucq, & J.D.R. Millán, "Decoding fast-paced error-related potentials in monitoring protocols." In Engineering in Medicine and Biology Society (EMBC), 2015 37th Annual International Conference of the IEEE (pp. 1111-1114). IEEE.

[13] J. Omedes, I. Iturrate, J. Minguez, & L. Montesano, "Analysis and asynchronous detection of gradually unfolding errors during monitoring tasks." Journal of neural engineering, 2015, 12(5), 056001.

[14] R. Chavarriaga, I. Iturrate, & J.D.R. Millán, "Robust, accurate spelling based on error-related potentials." In Proceedings of the 6th International Brain-Computer Interface Meeting 2016 (No. EPFL-CONF-218930).

[15] M. Völker, S. Berberich, E. Andreev, L.D.J. Fiederer, W. Burgard, & T. Ball., "Between-subject transfer learning for classification of error-related signals in high-density EEG." The First Biannual Neuroadaptive Technology Conference. Vol. 81. No. 8.8. 2017.

[16] M. Völker, L.D.J. Fiederer, S. Berberich, J. Hammer, J. Behncke, P. Kršek, M. Tomášek, P. Marusič, P. C. Reinacher, V.A. Coenen, M. Helias, A. Schulze-Bonhage, W. Burgard, & T.Ball, "The Dynamics of Error Processing in the Human Brain as Reflected by High-Gamma Activity in Noninvasive and Intracranial EEG." bioRxiv (2017): 166280.

[17] F. Burget*, L.D.J. Fiederer*, D. Kuhner*, D., M. Völker*, J. Aldinger*, R.T. Schirrmeister, C. Do, J. Boedecker, B. Nebel, T. Ball, & W. Burgard, "Acting Thoughts: Towards a mobile robotic service assistant for users with limited communication skills." 2017 arXiv preprint arXiv:1707.06633. * These authors contributed equally to the work.

[18] O. Ledoit & M. Wolf, "A well-conditioned estimator for large-dimensional covariance matrices." Journal of multivariate analysis 88.2 (2004): 365-411.

[19] M. Falkenstein, J. Hohnsbein, J. Hoormann, & L. Blanke, "Effects of crossmodal divided attention on late ERP components. II. Error processing in choice reaction tasks." Electroencephalography and clinical neurophysiology (1991), 78(6), 447-455.